%% file: acl2016.tex
\pdfoutput=1

\documentclass[11pt,a4paper]{article}
\usepackage{acl2016}
\usepackage{times}
\usepackage{url}
\usepackage{latexsym}
\usepackage{amsmath}

\usepackage{graphicx}
\usepackage{flexisym}
\usepackage{subfig}
\usepackage{helvet}
\usepackage{courier}
\usepackage{pgfplots}
\usepackage{dcolumn}
\usetikzlibrary{matrix,positioning}
\usepgfplotslibrary{groupplots}
\pgfplotsset{compat=newest}
\aclfinalcopy 
\title{Harnessing Cognitive Features for Sarcasm Detection}
\author{
\textbf{Abhijit Mishra}\textsuperscript{$\dagger$}, \textbf{Diptesh Kanojia}\textsuperscript{$\dagger$,$\clubsuit$}, \textbf{Seema Nagar} 
\textsuperscript{$\star$}, \textbf{Kuntal Dey}\textsuperscript{$\star$}, \\\textbf{Pushpak Bhattacharyya}\textsuperscript{$\dagger$}\\
\textsuperscript{$\dagger$}Indian Institute of Technology Bombay, India\\
\textsuperscript{$\clubsuit$}IITB-Monash Research Academy, India\\
\textsuperscript{$\star$}IBM Research, India\\
\textsuperscript{$\dagger$}\{abhijitmishra, diptesh, pb\}@cse.iitb.ac.in\\
\textsuperscript{$\star$}\{senagar3, kuntadey\}@in.ibm.com
}
\date{}
\begin{document}
\maketitle
\begin{abstract}
In this paper, we propose a novel mechanism for enriching the feature vector, for the task of sarcasm detection, with cognitive features extracted from eye-movement patterns of human readers. Sarcasm detection has been a challenging research problem, and its importance for NLP applications such as review summarization, dialog systems and sentiment analysis is well recognized. Sarcasm can often be traced to \textit{incongruity} that becomes apparent as the full sentence unfolds. This presence of incongruity- implicit or explicit- affects the way readers’ eyes move through the text. We observe the difference in the behaviour of the eye, while reading sarcastic and non sarcastic sentences. Motivated by this observation, we augment traditional linguistic and stylistic features for sarcasm detection with the cognitive features obtained from readers’ eye movement data. We perform statistical classification using the enhanced feature set so obtained. The augmented cognitive features improve sarcasm detection by $3.7\%$ (in terms of F-score), over the performance of the best reported system.
\end{abstract}

\section{Introduction}
\label{sec:intro}
Sarcasm is an intensive, indirect and complex construct that is often intended to express contempt or ridicule \footnote{The Free Dictionary}. Sarcasm, in speech, is multi-modal, involving tone, body-language and gestures along with linguistic artifacts used in speech.
Sarcasm in text, on the other hand, is more restrictive when it comes to such non-linguistic modalities. This makes recognizing textual sarcasm more challenging for both humans and machines.

Sarcasm detection plays an indispensable role in applications like online review summarizers, dialog systems, recommendation systems and sentiment analyzers. This makes automatic detection of sarcasm an important problem. However, it has been quite difficult 
to solve such a problem with traditional NLP tools and techniques. This is apparent from the results reported by the survey from \newcite{DBLP:journals/corr/JoshiBC16}. The following discussion brings more insights into this.

Consider a scenario where an online reviewer gives a negative opinion about a movie through sarcasm: \emph{``This is the kind of movie you see because the theater has air conditioning''}.
It is difficult for an automatic sentiment analyzer to assign a rating to the movie and, in the absence of any other information, such a system may not be able to comprehend that \emph{prioritizing the air-conditioning facilities of the theater over the movie experience indicates a negative sentiment towards the movie}.
This gives an intuition to why, for sarcasm detection, it is necessary to go beyond textual analysis.

We aim to address this problem by exploiting the psycholinguistic side of sarcasm detection, using cognitive features extracted with the help of \emph{eye-tracking}.
A motivation to consider cognitive features comes from analyzing human eye-movement trajectories that supports the conjecture: \textit{Reading sarcastic texts induces distinctive eye movement patterns, compared to literal texts.}
The cognitive features, derived from human eye movement patterns observed during reading, include two primary feature types:

\begin{enumerate}
\item Eye movement characteristic features of readers while reading given text, comprising \emph{gaze-fixaions} (\textit{i.e,}longer stay of gaze on a visual object), forward and backward \emph{saccades} (\emph{i.e.,} quick jumping of gaze between two positions of rest). 
\item Features constructed using the statistical and deeper structural information contained in \emph{graph}, created by treating words as vertices and saccades between a pair of words as edges.
\end{enumerate}

The cognitive features, along with textual features used in best available sarcasm detectors, are used to train binary classifiers against given sarcasm labels.
Our experiments show significant improvement in classification accuracy over the state of the art, by performing such augmentation.

\subsubsection{Feasibility of Our Approach}
\label{sec:feasibility}
Since our method requires gaze data from human readers to be available, the methods practicability becomes questionable. We present our views on this below.
\subsubsection{\textit{Availability of Mobile Eye-trackers}}
Availability of inexpensive embedded eye-trackers on hand-held devices has come close to reality now. This opens avenues to get eye-tracking data from inexpensive mobile devices from a huge population of online readers non-intrusively, and derive cognitive features to be used in predictive frameworks like ours.
For instance, \emph{Cogisen: (http://www.sencogi.com)} has a patent (ID: EP2833308-A1) on ``eye-tracking using inexpensive mobile web-cams". 
\subsubsection{\textit{Applicability  Scenario}}
We believe, mobile eye-tracking modules could be a part of mobile applications built for e-commerce, online learning, gaming \emph{etc.} where automatic analysis of online reviews calls for better solutions to detect linguistic nuances like sarcasm. To give an example, let's say a book gets different reviews on Amazon. Our system could watch how readers read the review using mobile eye-trackers, and thereby, decide  whether the text contains sarcasm or not. Such an application can horizontally scale across the web and will help in improving automatic classification of online reviews.

Since our approach seeks human mediation, one might be tempted to question the approach of relying upon eye-tracking, an indirect indicator, instead of directly obtaining man-made annotations. We believe, asking a large number of internet audience to annotate/give feedback on each and every sentence that they read online, following a set of annotation instructions, will be extremely intrusive and may not be responded well. Our system, on the other hand, can be seamlessly integrated into existing applications and as the eye-tracking process runs in the  background, users will not be interrupted in the middle of the reading. This, thus, offers a more natural setting where human mediation can be availed without intervention. 

\subsubsection{\textit{Getting Users' Consent for Eye-tracking}}
Eye-tracking technology has already been utilized by leading mobile technology developers (like Samsung) to facilitate richer user experiences through services  like \emph{Smart-scroll} (where a user's eye movement determines whether a page has to be scrolled or not) and \emph{Smart-lock} (where user's gaze position decides whether to lock the screen or not). The growing interest of users in using such services takes us to a promising situation where getting users' consent to record eye-movement patterns will not be difficult, though it is yet not the current state of affairs.

\textbf{Disclaimer: }In this work, we focus on detecting sarcasm in \emph{non-contextual} and \emph{short-text} settings prevalent in product reviews and social media. Moreover, our method requires eye-tracking data to be available in the test scenario.

\section{Related Work}
\label{sec:relatedwork}
Sarcasm, in general, has been the focus of research for quite some time. In one of the pioneering works \newcite{jorgensen1984test} explained how sarcasm arises when a figurative meaning is used opposite to the literal meaning of the utterance.
In the word of \newcite{clark1984pretense}, sarcasm processing involves canceling the indirectly negated message and replacing it with the implicated one. \newcite{giora1995irony}, on the other hand, define sarcasm as a  mode of indirect negation that requires processing of both negated and implicated messages. 
\newcite{ivanko2003context} define sarcasm as a six tuple entity consisting of \emph{a speaker, a listener, Context, Utterance, Literal Proposition} and \emph{Intended Proposition} and study the cognitive aspects of sarcasm processing.

Computational linguists have previously addressed this problem using rule based and statistical techniques, that make use of : \textbf{(a)} Unigrams and Pragmatic features \cite{carvalho2009clues,gonzalez2011identifying,barbieri2014modelling,joshi2015harnessing} \textbf{(b)} Stylistic patterns \cite{davidov2010semi} and patterns related to \emph{situational disparity}  \cite{riloff2013sarcasm} and \textbf{(c)} Hastag interpretations \cite{liebrecht2013perfect,maynard2014cares}.

Most of the previously done work on sarcasm detection uses \emph{distant supervision} based techniques (ex: leveraging \emph{hashtags}) and stylistic/pragmatic features (emoticons, laughter expressions such as \emph{``lol''} {\it etc}). But, detecting sarcasm in linguistically well-formed structures, in absence of explicit cues or information (like emoticons), proves to be  hard using such linguistic/stylistic features alone.

With the advent of sophisticated eye-trackers and electro/magneto-encephalographic (EEG/MEG) devices, it has been possible to delve deep into the cognitive underpinnings of sarcasm understanding.
\newcite{Filik2014}, using a series of eye-tracking and EEG experiments try to show that for unfamiliar ironies, the literal interpretation would be computed first. They also show that a mismatch with context would lead to a re-interpretation of the statement, as being ironic.
\newcite{Camblin2007103} show that in multi-sentence passages, discourse congruence has robust effects on eye movements. This also implies that disrupted processing occurs for discourse incongruent words, even though they are perfectly congruous at the sentence level. In our previous work \cite{sarcasmunderstandability}, we augment cognitive features, derived from eye-movement patterns of readers, with textual features to detect whether a human reader has realized the presence of sarcasm in text or not.  

The recent advancements in the literature discussed above, motivate us to explore gaze-based cognition for sarcasm detection. As far as we know, our work is the first of its kind.
\section{Eye-tracking Database for Sarcasm Analysis}
\label{sec:eyetracking}
Sarcasm often emanates from \emph{incongruity}~\cite{campbell2012there}, which enforces the brain to reanalyze 
it \cite{kutas1980reading}. This, in turn, affects the way eyes move through the text. Hence, \textbf{distinctive 
eye-movement patterns may be observed in the case of successful processing of 
sarcasm in text in contrast to literal texts}. This hypothesis forms the crux of 
our method for sarcasm detection and we validate this using our previously released freely available sarcasm dataset\footnote{\url{http://www.cfilt.iitb.ac.in/cognitive-nlp}} \cite{sarcasmunderstandability} enriched with gaze information.
\subsection{Document Description}
\label{subsec:eyetrackingdocdesc}
The database consists of 1,000 short texts, each having 10-40 words. Out of these, 350 are sarcastic and are collected as follows:
(a) 103 sentences are from two popular sarcastic quote 
websites\footnote{\url{http://www.sarcasmsociety.com},\\
\url{http://www.themarysue.com/funny-amazon-reviews}},
(b) 76 sarcastic short movie reviews are manually extracted from the 
\emph{Amazon Movie Corpus} \cite{pang2004sentimental} by two linguists.
(c) 171 tweets are downloaded using the hashtag \emph{\#sarcasm} from Twitter.
The 650 non-sarcastic texts are either downloaded from Twitter or extracted 
from the Amazon Movie Review corpus. The sentences do not contain words/phrases 
that are \emph{highly} topic or culture specific.
The tweets were normalized to make them linguistically well formed to avoid 
difficulty in interpreting social media lingo. Every sentence in our dataset 
carries positive or negative opinion about specific ``aspects''. For 
example, the sentence \emph{``The movie is extremely well cast''} has positive 
sentiment about the aspect ``cast''.

The annotators were seven graduate students with science and engineering background, and possess good English proficiency. They were given a set of instructions beforehand and are advised to seek clarifications before they proceed. The instructions mention the nature of the task, annotation input method, and necessity of head movement minimization during the experiment. 
\begin{table}[t]
\footnotesize
\begin{center}
\begin{tabular}{ c|c c c c c c }
\hline \hline
  & $\mu\_S$ & $\sigma\_S$ & $\mu\_NS$ & $\sigma\_NS$ & t & p \\
\hline
P1 & 319 & 145 & 196 & 97 & 14.1& 5.84E-39 \\
P2 & 415 & 192 & 253 & 130 & 14.0 & 1.71E-38\\
P3 & 322 & 173 & 214 & 160 & 9.5 & 3.74E-20\\
P4 & 328 & 170 & 191 & 96 & 13.9 & 1.89E-37\\
P5 & 291 & 151 & 183 & 76 & 11.9 & 2.75E-28\\
P6 & 230 & 118 & 136 & 84 & 13.2 & 6.79E-35\\
P7 & 488 & 268 & 252 & 141 & 15.3 & 3.96E-43\\
\hline
\end{tabular}
\end{center}
\caption{T-test statistics for average fixation duration time per word (in ms) for presence of sarcasm (represented by \emph{S}) and its absence (\emph{NS}) for participants P1-P7.}
\label{tab:ttest}
\end{table}

\subsection{Task Description}
The task assigned to annotators was to read sentences one at a time and label them with 
with binary labels indicating the polarity (\textit{i.e.,} positive/negative). Note that, the 
participants were not instructed to annotate whether a sentence is sarcastic or not., to rule out 
the \emph{Priming Effect} (\emph{i.e.,} if sarcasm is expected beforehand, processing incongruity becomes relatively easier
\cite{GibbsJr198641}). The setup ensures its ``ecological validity'' in two ways:
\textit{(1)} Readers are not given any clue that they have to treat sarcasm with special 
attention. This is done by setting the task to polarity annotation (instead of 
sarcasm detection). \textit{(2)} Sarcastic sentences are mixed with non sarcastic text, which does not give 
prior knowledge about whether the forthcoming text will be sarcastic or not.

The eye-tracking experiment is conducted by following the standard norms in eye-movement research \cite{holmqvist2011eye}.
At a time, one sentence is displayed to the reader along with the ``aspect'' with respect to which the annotation has to be provided.
While reading, an SR-Research Eyelink-1000 eye-tracker (monocular remote mode, sampling rate 500Hz) records several eye-movement parameters like fixations (a long stay of gaze) and saccade (quick jumping of gaze between two positions of rest) and pupil size.

The accuracy of polarity annotation varies between 72\%-91\% for sarcastic texts and 75\%-91\% for non-sarcastic text, showing the inherent difficulty of sentiment annotation, when sarcasm is present in the text under consideration.
Annotation errors may be attributed to: (a) lack of patience/attention while reading, (b) issues related to text comprehension, and (c) confusion/indecisiveness caused due to lack of context.

For our analysis, we do not discard the incorrect annotations present in the database. Since our system eventually aims to involve online readers for sarcasm detection, it will be hard to segregate readers who misinterpret the text. We make a rational assumption that, for a particular text, most of the readers, from a fairly large population, will be able to identify sarcasm. 
Under this assumption, the eye-movement parameters, averaged across all readers in our setting, may not be significantly distorted by a few readers who would have failed to identify sarcasm. This assumption is applicable for both  regular and multi-instance based classifiers explained in section \ref{sec:predictiveframework}.

\section{Analysis of Eye-movement Data}
\label{sec:analysis}
We observe distinct behavior during sarcasm reading, by analyzing the ``fixation duration on the text'' (also referred to as ``dwell time'' in the literature) and ``scanpaths'' of the readers.

\subsection{Variation in the  Average  Fixation Duration per Word}
Since sarcasm in text can be expected to induce cognitive load, it is reasonable to believe that it would require more processing time \cite{ivanko2003context}. Hence, fixation duration normalized over total word count should usually be higher for a sarcastic text than for a non-sarcastic one. We observe this for all participants in our dataset, with the \textit{average  fixation duration per word} for sarcastic texts being at least 1.5 times more than that of non-sarcastic texts.
To test the statistical significance, we conduct a two-tailed t-test (assuming unequal variance) to compare the average fixation duration per word  for sarcastic and non-sarcastic texts.
The hypothesized mean difference is set to 0 and the error tolerance limit ($\alpha$) is set to 0.05.
The t-test analysis, presented in Table~\ref{tab:ttest}, shows that for all participants, a statistically significant difference exists between the average fixation duration per word for sarcasm (higher average fixation duration) and non-sarcasm (lower average fixation duration). This affirms that the presence of sarcasm affects the duration of fixation on words.

\begin{figure}[t!]
\centering
\includegraphics[width=7.2cm,height=5cm]{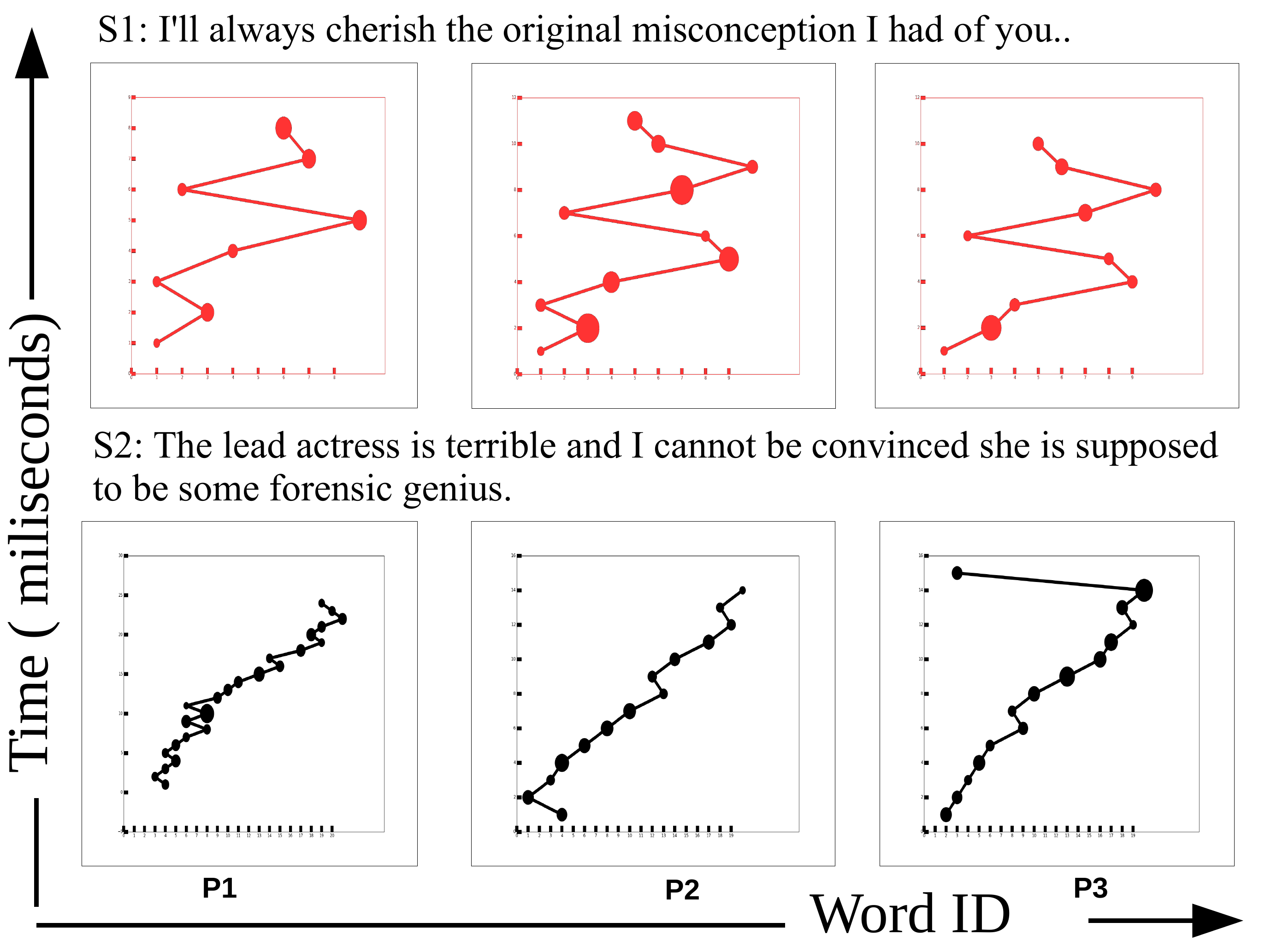}
\caption{Scanpaths of three participants for two negatively polar sentences sentence \emph{S1} and \emph{S2}. Sentence \emph{S1} is sarcastic but \emph{S2} is not.} 
\label{fig:sarcasmscanpath}
\vspace{-0.05in}
\end{figure}

It is important to note that longer fixations may also be caused by other linguistic subtleties (such as difficult words, ambiguity and syntactically complex structures) causing delay in comprehension, or occulomotor control problems forcing readers to spend time adjusting eye-muscles. So, an elevated average fixation duration per word may not sufficiently indicate the presence of sarcasm. But we would also like to share that, for our dataset, when we considered \textit{readability} (Flesch readability ease-score \cite{flesch1948new}), \textit{number of words in a sentence} and \textit{average character per word} along with the \textit{sarcasm label} as the predictors of average fixation duration following a linear mixed effect model \cite{barr2013random}, \textit{sarcasm label} turned out to be the most significant predictor with a maximum slope. This indicates that average fixation duration per word has a strong connection with the text being sarcastic, at least in our dataset.

We now analyze \emph{scanpaths} to gain more insights into the sarcasm comprehension process.
\begin{figure}[t]
\includegraphics[width=7.5cm,height=2.5cm]{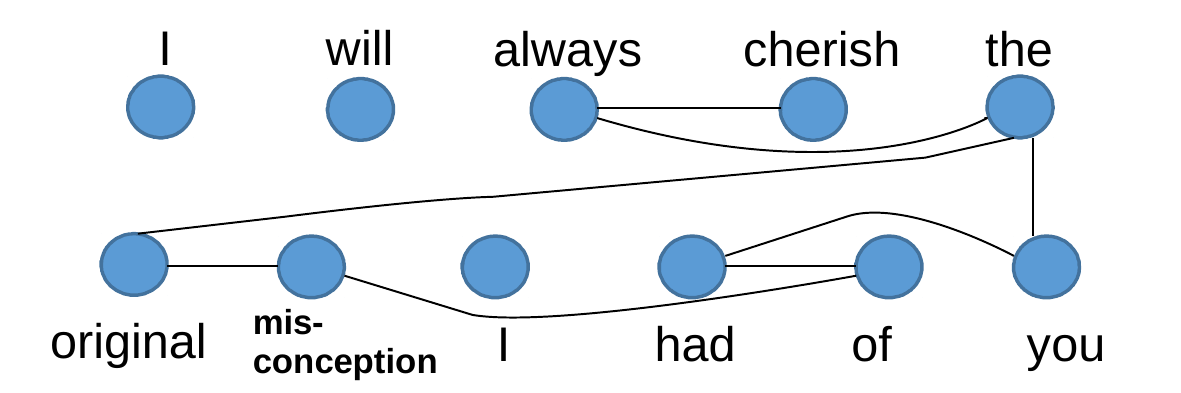}
\caption{Saliency graph of participant \emph{P1} for the sentence \emph{I will always cherish the original misconception I had of you.} } 
\label{fig:saliency}
\end{figure}

\subsection{Analysis of Scanpaths}
Scanpaths are line-graphs that contain fixations as nodes and saccades as edges; the radii of the nodes represent the fixation duration.
A scanpath corresponds to a participant's eye-movement pattern while reading a particular sentence.
Figure~\ref{fig:sarcasmscanpath} presents scanpaths of three  participants for the sarcastic sentence \textit{S1} and the non-sarcastic sentence \textit{S2}.
The x-axis of the graph represents the sequence of words a reader reads, and the y-axis represents a temporal sequence in milliseconds.  

Consider a sarcastic text containing incongruous phrases \emph{A} and \emph{B}.
Our qualitative scanpath-analysis reveals that scanpaths with respect to sarcasm processing have two typical characteristics.
Often, a long \emph{regression} - a saccade that goes to a previously visited segment - is observed when a reader starts reading \emph{B} after skimming through \emph{A}. In a few cases, the fixation duration on \emph{A} and \emph{B} are significantly higher than the average fixation duration per word.
In sentence \emph{S1}, we see long and multiple regressions from the two incongruous phrases \emph{``misconception''} and \emph{``cherish''}, and a few instances where phrases \emph{``always cherish''} and \emph{``original misconception''} are fixated longer than usual.
Such eye-movement behaviors are not seen for \emph{S2}.

Though sarcasm induces distinctive scanpaths like the ones depicted in Figure~\ref{fig:sarcasmscanpath} in the observed examples, presence of such patterns is not sufficient to guarantee sarcasm; such patterns may also possibly arise from literal texts.
We believe that a combination of linguistic features, readability of text and features derived from scanpaths would help discriminative machine learning models learn sarcasm better.
\input{featureTable.tex}
\begin{table*}[!th]
\centering
\footnotesize
\begin{tabular}{r||r r r|r r r|r r r|r}
\hline
\hline
\bf{Features} & \bf{P(1)} & \bf{P(-1)} & \bf{P(avg)} & \bf{R(1)} & \bf{R(-1)} & \bf{R(avg)} & \bf{F(1)} & \bf{F(-1)} & \bf{F(avg)} &\bf{Kappa}\\
\hline
\multicolumn{11}{c}{Multi Layered Neural Network}\\
\hline
Unigram  & 53.1 & 74.1 & 66.9 & 51.7 & 75.2 & 66.6 & 52.4 & 74.6 & 66.8 & 0.27 \\
Sarcasm (Joshi et. al.) & 59.2 & 75.4 & 69.7 & 51.7 & 80.6 & 70.4 & 55.2 & 77.9 & 69.9 & 0.33 \\
Gaze & 62.4 & 76.7 & 71.7 & 54 & 82.3 & 72.3 & 57.9 & 79.4 & 71.8 & 0.37 \\
Gaze+Sarcasm & 63.4 & 75 & 70.9 & 48 & 84.9 & 71.9 & 54.6 & 79.7 & 70.9 & 0.34 \\
\hline
\multicolumn{11}{c}{N{\"a}ive Bayes}\\
\hline
Unigram & 45.6 & 82.4 & 69.4 & 81.4 & 47.2 & 59.3 & 58.5 & 60 & 59.5 & 0.24 \\
Sarcasm (Joshi et. al.) & 46.1 & 81.6 & 69.1 & 79.4 & 49.5 & 60.1 & 58.3 & 61.6 & 60.5 & 0.25\\
Gaze & 57.3 & 82.7 & 73.8 & 72.9 & 70.5 & 71.3 & 64.2 & 76.1 & 71.9 & 0.41\\
Gaze+Sarcasm & 46.7 & 82.1 & 69.6 & 79.7 & 50.5 & 60.8 & 58.9 & 62.5 & 61.2 & 0.26\\
\hline
\multicolumn{11}{c}{Original system by Riloff et.al. : Rule Based with implicit incongruity}\\
\hline
Ordered & 60 & 30 & 49 & 50 & 39 & 46 & 54 & 34 & 47 &  0.10\\
\hline
Unordered & 56 & 28 & 46 & 40 & 42 & 41 & 46 & 33 & 42 & 0.16 \\
\hline
\multicolumn{11}{c}{Original system by Joshi et.al. : SVM with RBF Kernel}\\
\hline
Sarcasm (Joshi et. al.) & 73.1 & 69.4 & 70.7 & 22.6 & 95.5 & 69.8 & 34.5 & 80.4 & 64.2 & 0.21 \\
\hline
\multicolumn{11}{c}{SVM Linear: with default parameters}\\
\hline
Unigram & 56.5 & 77 & 69.8 & 58.6 & 75.5 & 69.5 & 57.5 & 76.2 & 69.6 & 0.34 \\
\emph{Sarcasm (Joshi et. al.)} & \emph{59.9} & \emph{78.7} & \emph{72.1} & \emph{61.4} & \emph{77.6} & \emph{71.9} & \emph{60.6} & \emph{78.2} & \emph{72} & \emph{0.39} \\
Gaze & \textbf{65.9} & 75.9 & 72.4 & 49.7 & 86 & 73.2 & 56.7 & 80.6 & 72.2 & 0.38 \\
\textbf{Gaze+Sarcasm} & 63.7 & 79.5 & 74 & 61.7 & 80.9 & 74.1 & 62.7 & 80.2 & 74 & 0.43 \\
\hline
\multicolumn{11}{c}{\textbf{Multi Instance Logistic Regression: Best Performing Classifier}}\\
\hline
Gaze & 65.3 & 77.2 & 73 & 53 & \textbf{84.9} & 73.8 & 58.5 & \textbf{80.8} & 73.1 & 0.41 \\
\textbf{Gaze+Sarcasm} & 62.5 & \textbf{84} & \textbf{76.5} & \textbf{72.6} & 76.7 & \textbf{75.3} & \textbf{67.2} & 80.2 & \textbf{75.7} & \textbf{0.47} \\
\hline
\end{tabular}
\caption{Classification results for different feature combinations. P$\rightarrow$ Precision, R$\rightarrow$ 
Recall, F$\rightarrow$ F_{score}, Kappa$\rightarrow$ Kappa statistics show \emph{agreement with the gold labels}. 
Subscripts 1 and -1 correspond to sarcasm and non-sarcasm classes respectively.}
\label{tab:featureCombinations}
\end{table*}
\section{Features for Sarcasm Detection}
\label{sec:features}
We describe the features used for sarcasm detection in Table~\ref{tab:featureset}. The features enlisted under \textit{lexical},\textit{implicit incongruity} and \textit{explicit incongruity} are borrowed from various literature (predominantly from \newcite{joshi2015harnessing}). These features are essential to separate sarcasm from other forms semantic incongruity in text (for example  ambiguity arising from \emph{semantic ambiguity} or from \emph{metaphors}).
Two additional \textit{textual} features \textit{viz.} \emph{readability} and \emph{word count} of the text are also taken under consideration.
These features are used to reduce the effect of text hardness and text length on the eye-movement patterns.

\subsection{Simple Gaze Based Features}
Readers' eye-movement behavior, characterized by fixations, forward saccades, skips and regressions, can be directly quantified by simple statistical aggregation (\textit{i.e.,} either computing features for individual participants and then averaging or performing a multi-instance based learning as explained in section \ref{sec:predictiveframework}). Since these eye-movement attributes relate to the cognitive process in reading  \cite{rayner1994eye}, we  consider these as features in our model. Some of these features have been reported by \newcite{sarcasmunderstandability} for modeling sarcasm understandability of readers. However, as far as we know, these features are being introduced in NLP tasks like textual sarcasm detection for the first time. The values of these features are believed to increase with the increase in the degree of surprisal caused by incongruity in text (except \emph{skip count}, which will decrease).
\subsection{Complex Gaze Based Features}
For these features, we rely on a graph structure, namely ``saliency graphs", derived from eye-gaze information and word sequences in the text. 
\subsubsection{Constructing Saliency Graphs:}
For each reader and each sentence, we construct a ``saliency graph'', representing the reader's attention characteristics.
A saliency graph for a sentence $S$ for a reader $R$, represented as $G= (V,E)$, is a graph with vertices ($V$) and edges ($E$) where each vertex $v \in V$ corresponds to a word in $S$ (may not be unique) and there exists an edge $e \in E$ between vertices $v_1$ and $v_2$ if R performs at least one saccade between the words corresponding to $v1$ and $v2$.

Figure~\ref{fig:saliency} shows an example of a saliency graph.
A saliency graph may be weighted, but not necessarily connected, for a given text (as there may be words in the given text with no fixation on them).
The ``complex'' gaze features derived from saliency graphs are also motivated by the theory of incongruity.
For instance, \emph{Edge Density} of a saliency graph increases with the number of distinct saccades, which could arise from the complexity caused by presence of sarcasm. 
Similarly, the highest weighted degree of a graph is expected to be higher, if the node corresponds to a phrase, incongruous to some other phrase in the text.

\begin{table*}[t]
\footnotesize
\begin{center}
\begin{tabular}{ c m{0.4cm} m{0.4cm} c c}
 \hline \hline
\textbf{Sentence} & \textbf{Gold} & \textbf{Sarcasm} & \textbf{Gaze} & \textbf{Gaze+Sarcasm} \\
\hline
\multicolumn{1}{m{10cm}|}{1. I would like to live in Manchester, England. The transition between Manchester and death would be unnoticeable.} & S & NS & S & S \\
\multicolumn{1}{m{10cm}|}{2. Helped me a lot with my panic attacks. I took 6 mg a day for almost 20 years. Can't stop of course but it makes me feel very comfortable.} & NS & S & NS & NS \\
\multicolumn{1}{m{10cm}|}{3. Forgot to bring my headphones to the gym this morning, the music they play in this gym pumps me up so much!} & S & S & NS & NS \\
\multicolumn{1}{m{10cm}|}{4. Best show on satellite radio!! No doubt about it. The little doggy company has nothing even close.} & NS & S & NS & S \\
 \hline
 \hline
\end{tabular}
\end{center}
\caption{Example test-cases with \emph{S} and \emph{NS} representing labels for sarcastic and not-sarcastic respectively.}
\label{tab:examples}
\end{table*}
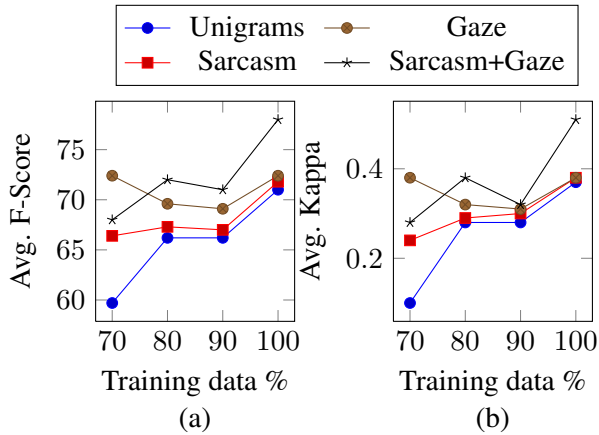
\begin{figure}[h]
\begin{tikzpicture}
\pgfplotsset{compat=newest}
 \begin{groupplot}[group style={group name=my plots, group size= 2 by 1,horizontal sep=1.3 cm },height=4.5cm,width=4.2cm]
\nextgroupplot[
    xlabel={Training data \%},
    ylabel={Avg. F-Score},
    legend style = { column sep = 5pt, legend columns = 2, transpose legend,legend to name = grouplegend1}]
]
\addplot coordinates {

      (70,59.7)    (80,66.2)   (90,66.2) (100,71)
      
};\addlegendentry{Unigrams}
\addplot coordinates {

      (70,66.4)    (80,67.3)   (90,67) (100,71.8)
      
};\addlegendentry{Sarcasm}
\addplot coordinates {

      (70,72.4)    (80,69.6)   (90,69.1) (100,72.4)
      
};\addlegendentry{Gaze}
\addplot coordinates {

      (70,68)    (80,72)   (90,71) (100,78)

};\addlegendentry{Sarcasm+Gaze}
\nextgroupplot[
    xlabel={Training data \%},
    ylabel={Avg. Kappa},
]
\addplot coordinates {

      (70,0.1)    (80,0.28)   (90,0.28) (100,0.37)
      
};\label{plot1:unigram};
\addplot coordinates {

      (70,0.24)    (80,0.29)   (90,0.30) (100,0.38)
      
};\label{plot1:sarcasm};
\addplot coordinates {

      (70,0.38)    (80,0.32)   (90,0.31) (100,0.38)
      
};\label{plot1:gaze};
\addplot coordinates {

      (70,0.28)    (80,0.38)   (90,0.32) (100,0.51)
      
};\label{plot1:gl};
\end{groupplot}
\node[below = 1cm of my plots c1r1.south] {(a)};
\node[below = 1cm of my plots c2r1.south] {(b)};
\node at ($(my plots c1r1) - (-2.0cm,-2.2cm)$) {\ref{grouplegend1}};
\end{tikzpicture}
\caption{Effect of training data size on classification in terms of (a) F-score and (b) \emph{Kappa} statistics}
\label{fig:VarTraining}
\end{figure}
\begin{figure}[!th]
\centering
\begin{tikzpicture}
\pgfplotsset{compat=newest}
\pgfplotsset{every y tick label/.append style={font=\tiny}}
 \begin{groupplot}[group style={group name=my plots2, group size= 1 by 2,vertical sep=1 cm },height=6cm,width=6cm]
\nextgroupplot[
    xbar,enlargelimits=0.1,
    ylabel={Y: 	Features},
    xlabel={X: Avg. Merit (Chi-squared)},
    ytick = {1,2,3,4,5,6,7,8,9,10,11,12,13,14,15,16,17,18,19,20},
    yticklabels = {RED,*RDSH*,*ED*,*FC*,*PSS*,*RSH*,*PSH*,+VE,*F1S*,*SKIP*,*SL*,*F1H*,*RSS*,IMP,*F2S*,UNI,LEN,*F2H*,*LREG*,*FDUR*},
    extra y ticks={2,3,4,5,6,7,9,10,11,12,13,15,18,19,20},
    extra y tick labels = {*RDSH*,*ED*,*FC*,*PSS*,*RSH*,*PSH*,*F1S*,*SKIP*,*SL*,*F1H*,*RSS*,*F2S*,*F2H*,*LREG*,*FDUR*},
    extra y tick style={
      yticklabel style={color=brown,font=\tiny}
    },
    bar width=3pt
]
\addplot [draw=blue,fill=blue!25]
coordinates{
(94.801,20) (60.02,19) (51.7415,18) (49.635,17) (49.811,16) (49.2165,15) (47.415,14) (39.0585,13) (36.3575,12) (36.0955,11) (33.4665,10) (32.31,9) (31.084,8) (31.663,7) (30.3435,6) (27.585,5) (26.964,4) (22.954,3) (17.444,2) (17.404,1)
};
\nextgroupplot[
    xbar,enlargelimits=0.1,
    ylabel={Y: Features},
    xlabel={X: Avg. Merit (InfoGain)},
    ytick = {1,2,3,4,5,6,7,8,9,10,11,12,13,14,15,16,17,18,19,20},
    yticklabels = {*RSDH*,*RSDS*,*ED*,*PSS*,*FC*,*RSH*,*F1S*,*PSH*,*SKIP*,+VE,*SL*,*RSS*,*F1H*,*F2S*,IMP,*F2H*,UNI,LEN,*LREG*,*FDUR*},
    extra y ticks={1,2,3,4,5,6,7,8,9,11,12,13,14,16,19,20},
    extra y tick labels = {*RSDH*,*RSDS*,*ED*,*PSS*,*FC*,*RSH*,*F1S*,*PSH*,*SKIP*,*SL*,*RSS*,*F1H*,*F2S*,*F2H*,*LREG*,*FDUR*},
    extra y tick style={
      yticklabel style={color=brown,font=\tiny}
    },
    bar width=3pt
]
\addplot [draw=red,fill=red!25]
coordinates{
(85.5,20) (55.5,19) (45.5,18) (45,17) (44,16) (42,15) (41.5,14) (33.5,13) (33,12) (31.5,11) (30,10) (28,9) (27,8) (26.5,7) (25,6) (24,5) (23,4) (19.5,3) (17,2) (15.5,1)
};
\end{groupplot}
\end{tikzpicture}
\caption{Significance of features observed by ranking the features using \textit{Attribute Evaluation based on Information Gain} and \textit{Attribute Evaluation based on Chi-squared test}. The length of the bar corresponds to the average merit of the feature. Features marked with * are gaze features.}
\label{fig:Significance}
\end{figure}
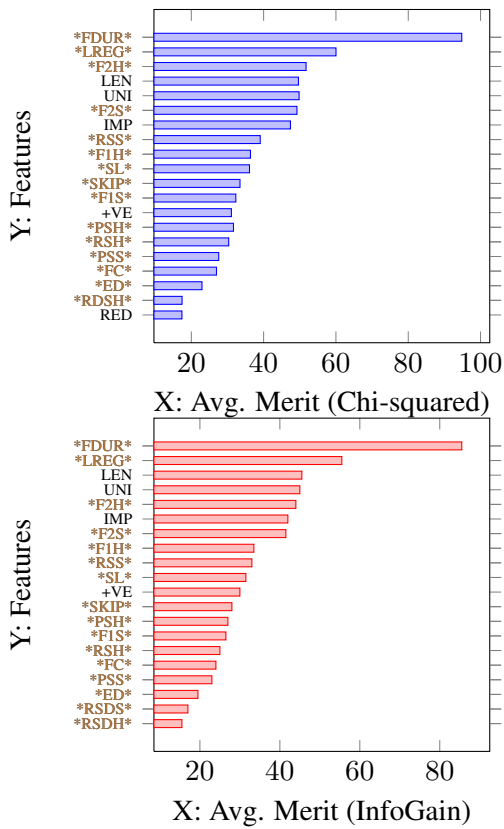

\section{The Sarcasm Classifier}
\label{sec:predictiveframework}
We interpret sarcasm detection as a binary classification problem.
The training data constitutes 994 examples created using our eye-movement database for sarcasm detection.
To check the effectiveness of our feature set, we observe the performance of multiple classification techniques on our dataset through a \emph{stratified} 10-fold cross validation.
We also compare the classification accuracy of our system and the best available systems proposed by \newcite{riloff2013sarcasm} and \newcite{joshi2015harnessing} on our dataset.
Using Weka ~\cite{hall2009weka} and LibSVM~\cite{libsvm2011} APIs, we implement the following classifiers:

\begin{itemize}
\item N{\"a}ive Bayes classifier
\item Support Vector Machines~\cite{cortes1995support} with default hyper-paramaters
\item Multilayer Feed Forward Neural Network
\item Multi Instance Logistic Regression (MILR)~\cite{xu2004logistic}
\end{itemize}

\subsection{Results}
Table \ref{tab:featureCombinations} shows the classification results considering various feature combinations for different classifiers and other systems.
These are:
\begin{itemize}
\item \textit{Unigram} (with principal components of unigram feature vectors),
\item \textit{Sarcasm} (the feature-set reported by \newcite{joshi2015harnessing} subsuming unigram features and features from other reported systems)
\item \textit{Gaze} (the simple and complex cognitive features we introduce, along with readability and word count features), and
\item \textit{Gaze+Sarcasm} (the complete set of features).
\end{itemize}

For all regular classifiers, the gaze features are averaged across participants and augmented with linguistic and sarcasm related features. For the MILR classifier, the gaze features derived from each participant are augmented with linguistic features and thus, a multi instance ``bag'' of features is formed for each sentence in the training data. This multi-instance dataset is given to an MILR classifier, which follows the \emph{standard multi instance assumption} to derive class-labels for each bag.

For all the classifiers, our feature combination outperforms the baselines (considering only unigram features) as well as~\cite{joshi2015harnessing}, with the MILR classifier getting an F-score improvement of  \textbf{3.7\%} and \emph{Kappa} difference of \textbf{0.08}. We also achieve an improvement of  \textbf{2\%} over the baseline, using SVM classifier, when we employ our feature set. We also observe that the gaze features alone, also capture the differences between sarcasm and non-sarcasm classes with a high-precision but a low recall. 

To see if the improvement obtained is statistically significant over the state-of-the art system with textual sarcasm features alone, we perform  \textbf{McNemar test}.
The output of the SVM classifier using only linguistic features used for sarcasm detection by \newcite{joshi2015harnessing} and the output of the MILR classifier with the complete set of features are compared, setting threshold $\alpha = 0.05$.
There was a significant difference in the classifier's accuracy with \textbf{p(two-tailed) = 0.02} with an odds-ratio of \textbf{1.43}, showing that the classification accuracy improvement is unlikely to be observed by chance in 95\% confidence interval.
\subsection{Considering Reading Time as a Cognitive Feature along with Sarcasm Features}
One may argue that, considering simple measures of reading effort like ``reading time'' as cognitive feature instead of the expensive eye-tracking features for sarcasm detection may be a cost-effective solution. To examine this, we repeated our experiments with ``reading time'' considered as the only cognitive feature, augmented with the textual features. The F-scores of all the classifiers turn out to be close to that of the classifiers considering sarcasm feature alone and the difference in the improvement is not statistically significant ($p > 0.05$). One the other hand, F-scores with gaze features are superior to the F-scores when  reading time is considered as a  cognitive feature.
\subsection{How Effective are the Cognitive Features}
We examine the effectiveness of cognitive features on the classification accuracy by varying the input training data size.
To examine this, we create a stratified (keeping the class ratio constant) random train-test split of 80\%:20\%.
We train our classifier with 100\%, 90\%, 80\% and 70\% of the training data with our whole feature set, and the feature combination from \newcite{joshi2015harnessing}.
The goodness of our system is demonstrated by improvements in F-score and Kappa statistics, shown in Figure~\ref{fig:VarTraining}.

We further analyze the importance of features by ranking the features based on \textbf{(a)} Chi squared test, and \textbf{(b)} Information Gain test, using Weka's attribute selection module.
Figure~\ref{fig:Significance} shows the top 20 ranked features produced by both the tests.
For both the cases, we observe 16 out of top 20 features to be gaze features.
Further, in each of the cases, \emph{Average Fixation Duration per Word} and \emph{Largest Regression Position} are seen to be the two most significant features.

\subsection{Example Cases}
Table \ref{tab:examples} shows a few example cases from the experiment with stratified 80\%-20\% train-test split.

\begin{itemize}
\item Example sentence 1 is  sarcastic, and requires extra-linguistic knowledge (about poor living conditions at Manchester). Hence, the sarcasm detector relying only on textual features is unable to detect the underlying incongruity. However, our system predicts the label successfully, possibly helped by the gaze features.
\item Similarly, for sentence 2, the false sense of presence of incongruity (due to phrases like ``Helped me'' and ``Can't stop'') affects the system with only linguistic features. Our system, though, performs well in this case also.
\item Sentence 3 presents a false-negative case where it was hard for even humans to get the sarcasm. This is why our gaze features (and subsequently the complete set of features) account for erroneous prediction.
\item In sentence 4, gaze features alone false-indicate presence of incongruity, whereas the system predicts correctly when gaze and linguistic features are taken together.
\end{itemize}

From these examples, it can be inferred that, only gaze features would not have sufficed to rule out the possibility of detecting other forms of incongruity that do not result in sarcasm.

\subsection{Error Analysis}
Errors committed by our system arise from multiple factors, starting from limitations of the eye-tracker hardware to errors committed by linguistic tools and resources.
Also, aggregating various eye-tracking parameters to extract the cognitive features may have caused information loss in the regular classification setting.
\section{Conclusion}
\label{sec:conclusion}
In the current work, we created a novel framework to detect sarcasm, that derives insights from human cognition, that manifests over eye movement patterns.
We hypothesized that distinctive eye-movement patterns, associated with reading sarcastic text, enables improved detection of sarcasm.
We augmented traditional linguistic features with cognitive features obtained from readers' eye-movement data in the form of simple gaze-based features and complex features derived from a graph structure.
This extended feature-set improved the success rate of the sarcasm detector by 3.7\%, over the best available system.
Using cognitive features in an NLP Processing system like ours is the first proposal of its kind.

Our general approach may be useful in other NLP sub-areas like sentiment and emotion analysis, text summarization and question answering, where considering textual clues alone does not prove to be sufficient.
We propose to augment this work in future by exploring deeper graph and gaze features.
We also propose to develop models for the purpose of learning complex gaze feature representation, that accounts for the power of individual eye movement patterns along with the aggregated patterns of eye movements.
\section*{Acknowledgments}
We thank the members of CFILT Lab, especially Jaya Jha and Meghna Singh, and the students of IIT Bombay for their help and support. 
\bibliographystyle{acl2016}
\bibliography{acl2016.bib}
\end{document}

%% file: featureTable.tex
\begin{table*}[!th]
\centering
\footnotesize
\begin{tabular}{|p{1.5cm}|p{4.8cm}|p{0.8cm}|p{7cm}|}
\hline
\textbf{Subcategory} & \textbf{Feature Name} & \textbf{Type} & \textbf{Intent} \\
\hline
\multicolumn{4}{|c|}{{\it Category: Textual Sarcasm Features, Source: Joshi et. al.}} \\
\hline
Lexical & Presence of Unigrams (UNI) & Boolean & Unigrams in the training corpus \\
\cline{2-4}
& Punctuations (PUN) & Real & Count of punctuation marks \\
\hline
Implicit Incongruity & Implicit Incongruity (IMP) & Boolean & Incongruity of extracted implicit phrases (Rilof et.al, 2013) \\
\hline
& Explicit Incongruity (EXP) & Integer & Number of times a word follows a word of opposite polarity \\
\cline{2-4}
& Largest Pos/Neg Subsequence (LAR) & Integer & Length of the largest series of words with polarities unchanged \\
\cline{2-4}
Explicit & Positive words (+VE) & Integer & Number of positive words \\
\cline{2-4}
Incongruity & Negative words (-VE) & Integer & Number of negative words \\
\cline{2-4}
& Lexical Polarity (LP) & Integer & Sentence polarity found by supervised logistic regression \\
\hline
\multicolumn{4}{|c|}{{\it Category: Cognitive Features. We introduce these features for sarcasm detection.}} \\
\hline
& Readability (RED) & Real & Flesch Readability Ease \cite{flesch1948new} score of the sentence \\
\cline{2-4}
Textual & Number of Words (LEN) & Integer & Number of words in the sentence \\
\hline
& Avg. Fixation Duration (FDUR) & Real & Sum of fixation duration divided by word count \\
\cline{2-4}
& Avg. Fixation Count (FC) & Real & Sum of fixation counts divided by word count \\
\cline{2-4}
& Avg. Saccade Length (SL) & Real & Sum of saccade lengths (measured by number of words) divided by word count \\
\cline{2-4}
Simple & Regression Count (REG) & Real & Total number of gaze regressions \\
\cline{2-4}
Gaze & Skip count (SKIP) & Real & Number of words skipped divided by total word count \\
\cline{2-4}
Based & Count of regressions from second half to first half of the sentence (RSF) & Real & Number of regressions from second half of the sentence to the first half of the sentence (given the sentence is divided into two equal half of words) \\
\cline{2-4}
& Largest Regression Position (LREG) & Real & Ratio of the absolute position of the word from which a regression with the largest amplitude (number of pixels) is observed, to the total word count of sentence \\
\hline
& Edge density of the saliency gaze graph (ED) & Real & Ratio of the number of directed edges to vertices in the saliency gaze graph (SGG) \\
\cline{2-4}
& Fixation Duration at Left/Source (F1H, F1S) & Real & Largest weighted degree (LWD) and second largest weighted degree (SWD) of the SGG considering the fixation duration of word $i$ of edge $E_{ij}$ \\
\cline{2-4}
Complex & Fixation Duration at Right/Target (F2H, F2S) & Real & LWD and SWD of the SGG considering the fixation duration of word $j$ of edge $E_{ij}$ \\
\cline{2-4}
Gaze & Forward Saccade Word Count of Source (PSH, PSS) & Real & LWD and SWD of the SGG considering the number of forward saccades between words $i$ and $j$ of an edge $E_{ij}$ \\
\cline{2-4}
Based & Forward Saccade Word Count of Destination (PSDH, PSDS) & Real & LWD and SWD of the SGG considering the total distance (word count) of forward saccades between words $i$ and $j$ of an edge $E_{ij}$ \\
\cline{2-4}
& Regressive Saccade Word Count of Source (RSH, RSS) & Real & LWD and SWD of the SGG considering the number of regressive saccades between words $i$ and $j$ of an edge $E_{ij}$ \\
\cline{2-4}
& Regressive Saccade Word Count of Destination (RSDH, RSDS) & Real & LWD and SWD of the SGG considering the total distance (word count) of regressive saccades between words $i$ and $j$ of an edge $E_{ij}$ \\
\hline
\end{tabular}
\caption{The complete set of features used in our system.}
\label{tab:featureset}
\end{table*}